\documentclass[letterpaper, 10 pt, conference]{ieeeconf}
\IEEEoverridecommandlockouts
\usepackage{caption}
\usepackage[utf8]{inputenc} 

\usepackage[T1]{fontenc}    
\usepackage{hyperref}       
\usepackage{url}            
\usepackage{booktabs}       
\usepackage{amsfonts}       
\usepackage{nicefrac}       
\usepackage{microtype}      
\usepackage{xcolor}         
\usepackage{multirow}
\usepackage{graphicx}
\usepackage{makecell}

\usepackage[T1]{fontenc}
\usepackage{lmodern}  
\usepackage{graphicx}
\usepackage{amssymb}
\usepackage{booktabs}
\usepackage{multirow}
\usepackage{multicol}
\usepackage{caption} 
\usepackage{longtable} 
\usepackage{color}
\usepackage{pifont}%
\usepackage{array}
\usepackage{colortbl}

\usepackage{amsmath,amssymb,amsfonts} 
\usepackage{mathtools}                 
\usepackage{bm}                        
\usepackage{siunitx}                   
\usepackage{microtype}                 

\usepackage{newtxtext}                 

\begin{document}


\title{\LARGE \bf InsightDrive: Insight Scene Representation for End-to-End \\ Autonomous Driving}

\author{
    Ruiqi Song\textsuperscript{1,2,*}, 
    Xianda Guo\textsuperscript{3,*}, 
    Yanlun Peng\textsuperscript{4}, 
    Qinggong Wei\textsuperscript{5}, 
    Hangbin Wu\textsuperscript{1,\textdagger}, 
    Long Chen\textsuperscript{2,6,$^\dagger$} \\
    \textsuperscript{1} College of Surveying and Geo-informatics, Tongji University \\
    \textsuperscript{2} Institute of Automation, Chinese Academy of Sciences\\
~~~~\textsuperscript{3} School of Computer Science, Wuhan University \\~~~~\textsuperscript{5} The School of Artificial Intelligence, University of Chinese Academy of Sciences\\ ~~~~\textsuperscript{4} Great Wall Motor ~~~~\textsuperscript{6} IAIR, Xi'an Jiaotong University
\\
\texttt{ruiqi.song@ia.ac.cn;
xianda\_guo@163.com; yanlunpeng@gwm.cn;}\\
\texttt{weiqinggong20@mails.ucas.ac.cn; hb@tongji.edu.cn;
long.chen@ia.ac.cn}
\vspace{-5mm}
}


\maketitle

\renewcommand{\thefootnote}{\fnsymbol{footnote}}
\footnotetext[1]{These authors contributed equally to this work.}
\footnotetext[2]{Corresponding Author.}

\begin{abstract}
Conventional end-to-end autonomous driving methods often rely on explicit global scene representations, which typically consist of 3D object detection, online mapping, and motion prediction. In contrast, human drivers selectively attend to task-relevant regions and implicitly reason over the broader traffic context. Motivated by this observation, we introduce a lightweight end-to-end autonomous driving framework, InsightDrive. Unlike approaches that directly embed large language models (LLMs), InsightDrive introduces an Insight scene representation that jointly models attention-centric explicit scene representation and reasoning-centric implicit scene representation, so that scene understanding aligns more closely with human cognitive patterns for trajectory planning. To this end, we employ Chain-of-Thought (CoT) instructions to model human driving cognition and design a task-level Mixture-of-Experts (MoE) adapter that injects this knowledge into the autonomous driving model at negligible parameter cost. We further condition the planner on both explicit and implicit scene representations and employ a diffusion-based generative policy, which produces robust trajectory predictions and decisions. The overall framework establishes a knowledge distillation pipeline that transfers human driving knowledge to LLMs and subsequently to onboard models. Extensive experiments on the nuScenes and Navsim benchmarks demonstrate that InsightDrive achieves significant improvements over conventional scene representation approaches.

\end{abstract}

\section{Introduction}

Modular design, which decomposes autonomous driving systems into perception, prediction, planning, and control modules, has long been the mainstream paradigm due to its flexibility and interpretability. However, such architectures are heavily relying on manually designed rules, which limits their adaptability in complex scenarios. Recently, end-to-end autonomous driving methods that directly generate planning results from raw sensor inputs have attracted growing attention, as they demonstrate higher robustness and generalization. However, most existing approaches still depend on explicit 3D perception tasks (e.g., detection, segmentation, and occupancy prediction) as intermediate supervision to refine planning features. For instance, UniAD\cite{uniad} used dense prediction tasks, such as map segmentation, object detection, and occupancy prediction, as intermediate explicit outputs to refine the features of the planning task. VAD\cite{vad, vadv2} and GenAD\cite{genad} adopt vectorized scene representations, encoding maps and objects as vectors to improve inference efficiency. This design still emphasizes global scene representation, which may overlook planning-critical regions and lacks the capability for scene understanding and reasoning.

\begin{figure}[t]
    \centering
    \includegraphics[width=0.48\textwidth]{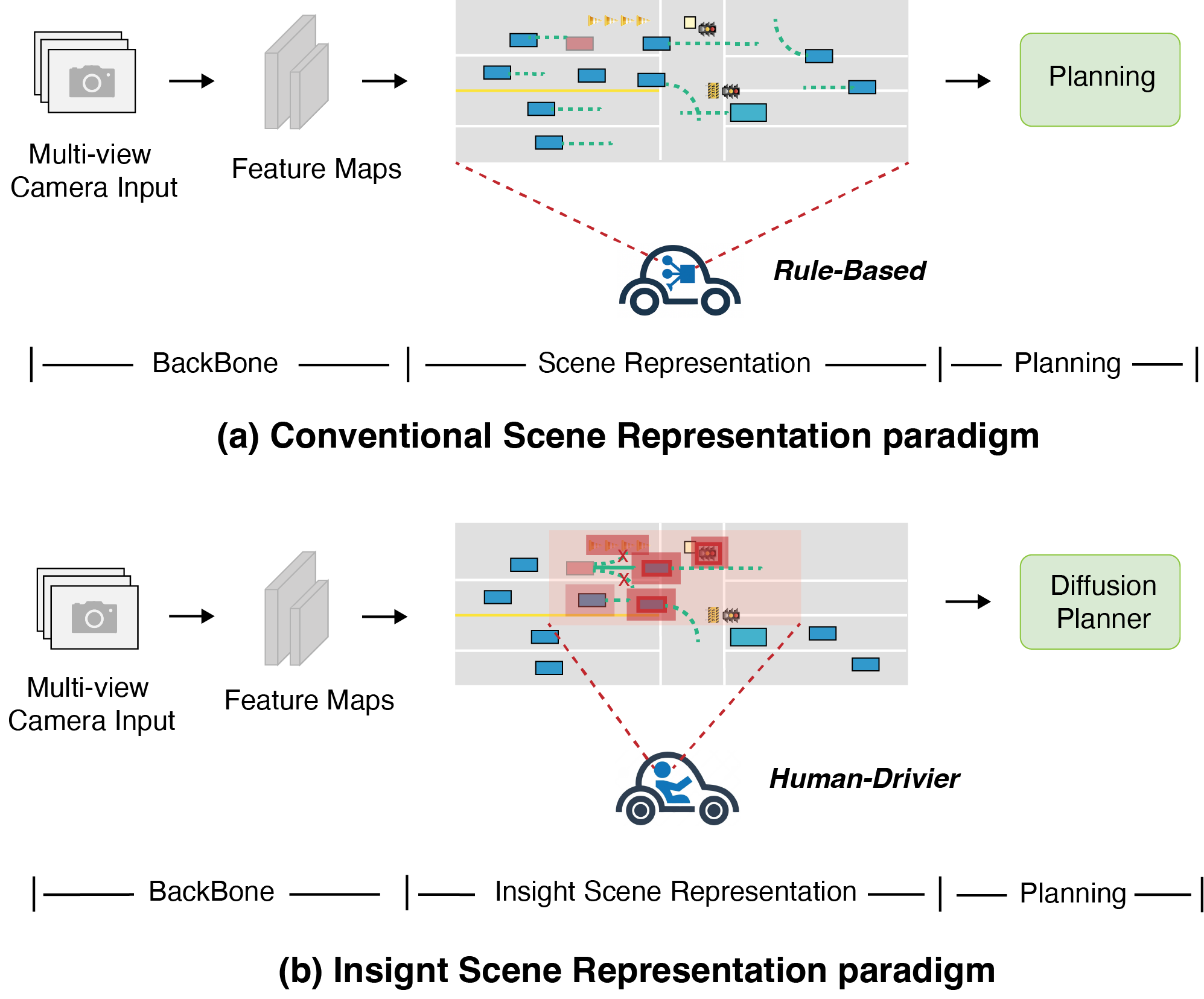} %
    \caption{Comparison between the proposed insight scene representation end-to-end autonomous driving framework with the conventional pipeline.}
    \label{fig1}
\end{figure}

To overcome these limitations, we propose InsightDrive, a lightweight end-to-end autonomous driving framework (illustrated in Fig.~\ref{fig1}), which jointly models attention-centric explicit scene representation and reasoning-based implicit scene representation, aligning scene understanding more closely with human driving cognition. Specifically, we construct Chain-of-Thought (CoT) instructions to simulate human cognitive processes and fine-tune a large language model (LLM) to generate driving knowledge. A task-level mixture-of-experts (MoE) adapter is then introduced to inject this knowledge into conventional scene representations, which allows the model to perform attention-centric perception and reasoning with negligible parameter overhead. Moreover, A diffusion planner conditioned on explicit and implicit scene representations is proposed to produce robust, adaptive trajectories. The overall pipeline establishes a knowledge distillation path from human drivers → LLMs → onboard models, which transfers human attention patterns and reasoning abilities into autonomous driving systems.

The main contributions of this work are as follows:
\textbf{1)} We present InsightDrive, which leverages CoT instructions to fine-tune LLMs and establishes a human–LLM–vehicle distillation pipeline that transfers human driving cognition into onboard models for joint explicit and implicit scene representation.
\textbf{2)} We design a Task-level Mixture-of-Experts adapter that injects human driving cognitive processes into scene representations with minimal parameter overhead, which enhances scene understanding and reasoning.
\textbf{3)} We propose a diffusion-based generative planner that uses explicit attention and implicit reasoning as conditions for generating robust and adaptive trajectories.
\textbf{4)} We conduct comprehensive experiments on both the nuScenes and Navsim benchmarks, which demonstrate the effectiveness and robustness of InsightDrive and show that it achieves state-of-the-art performance.
\section{Related work}
\subsection{Scene Representation in End-to-End Autonomous Driving}
Most prior end-to-end autonomous driving approaches sample explicit scene representation, which derived from traditional perception tasks, from dense BEV features based on Transformer framework. ST-P3 \cite{stp3} introduced the first end-to-end autonomous driving framework with explicit scene representation, which consists of object detection and BEV semantic segmentation, based on surround-view cameras. UniAD \cite{uniad}, building on prior work, integrated perception, tracking, prediction, and planning into a Transformer-based end-to-end framework, achieving state-of-the-art performance across all relevant tasks on the nuScenes\cite{nuscenes}. In addition to traditional 3D object detection and tracking, it introduces semantic occupancy prediction for explicit scene representation. VAD \cite{vad}, VADv2 \cite{vadv2}, and GenAD \cite{genad} vectorized the dense scene representations, simplifying perception tasks through vectorized BEV object detection and map segmentation, achieving better performance compared with UniAD\cite{uniad}. PARA-Drive \cite{paradrive} reorganized the cascaded framework into a parallel one, utilizing semantic maps and occupancy prediction for scene representation, while deactivating certain tasks during inference. OccWorld \cite{occworld} explored a dynamically evolving scene representation by learning a world model in 3D occupancy space, while predicting both the motion of the ego-vehicle and the evolution of the surrounding scene. These methods, which sample from dense BEV features as the primary scene, increase model complexity, especially for the dense occupancy prediction tasks\cite{scene}.

With the emergence of sparse scene representations, recent sparse end-to-end autonomous driving methods directly interact with task-specific queries and image features. SparseDrive \cite{SparseDriveEA}, SparseAD \cite{sparsead} and DiffusionDrive\cite{diffusiondrive} decouple instance features and geometric anchors from image features to fully represent instances, including dynamic road agents and static map elements. 
WoTE\cite{wote} uses a BEV world-model representation to encode scene dynamics, perform online rollouts to score candidate trajectories, and execute in closed loop the lowest-risk plan.
It is worth noting that the scene representations in end-to-end autonomous driving evolve from dense to sparse, and from sparse to implicit. However, the emphasis remains on global scene representation, lacking explicit guidance and attention to key instance tokens that impact driving, and lacking interpretability.

\subsection{Interpretability for End-to-End Autonomous Driving}
\begin{figure*}[t]
\centering 
\includegraphics[width=1\textwidth]{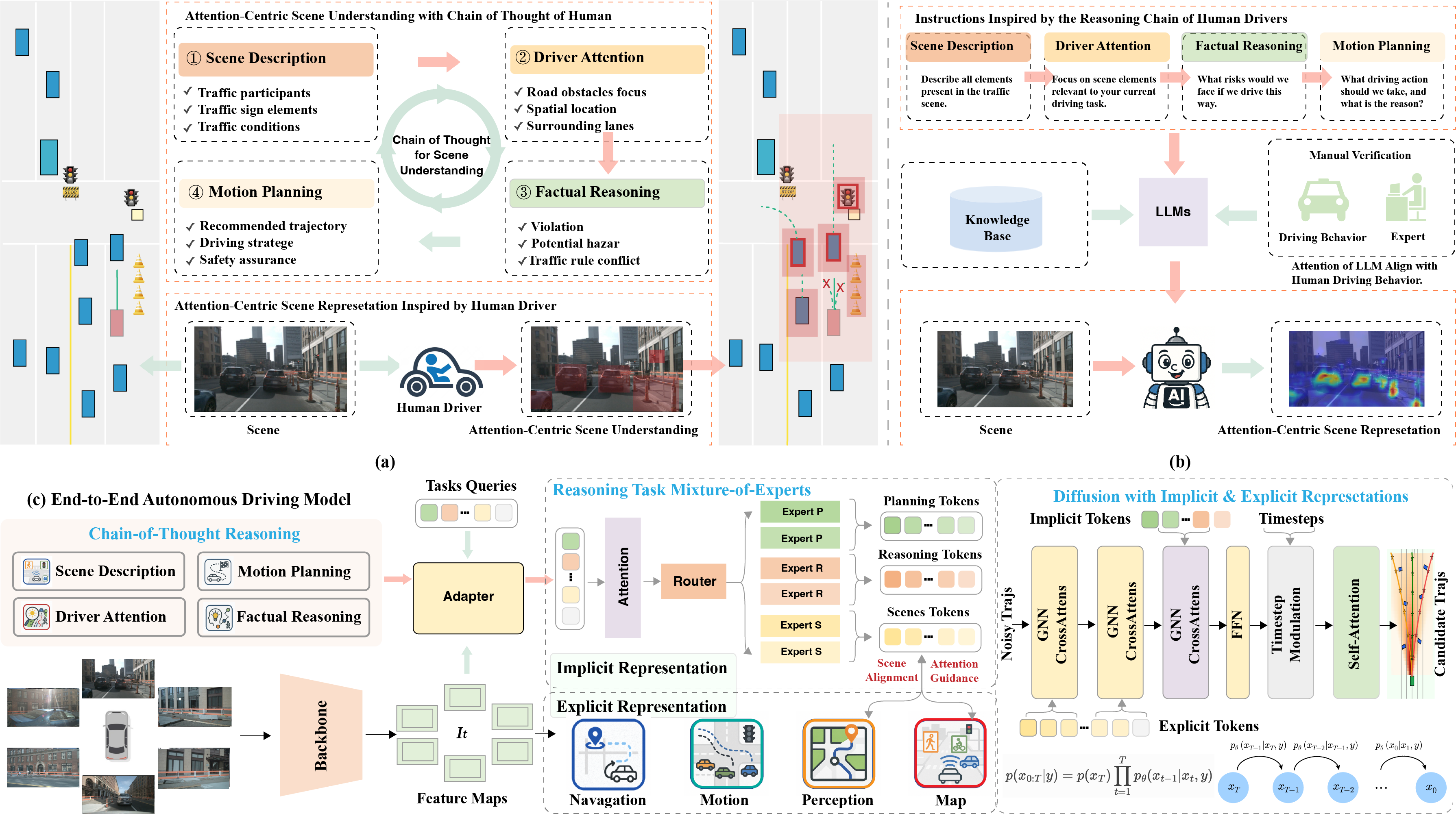}
\caption{\textbf{Framework of our insight scene representation for end-to-end autonomous driving.}
(a) Insight scene understanding with chain of thought of human.
(b) Instructions inspired by the reasoning chain of human drivers.
(c) Lightweight end-to-end autonomous driving model.
}
\label{fig:framework}
\end{figure*}

With the emergence of Large Language Models (LLMs) and Vision-Language Models (VLMs), research in autonomous driving has gradually shifted toward more efficient and interpretable architectures and started to integrate these models to enhance reasoning and decision-making capabilities\cite{drivevlm,gpt-driver,rda,opendrivevla}.
In Hint-AD \cite{hint}, a global alignment approach for interpretability is proposed to improve the transparency and traceability of decision-making in end-to-end models. DriveGPT4 \cite{drivegpt4} explore how to integrate large language models into end-to-end autonomous driving systems to enhance interpretability and reasoning capabilities. 

In addition, vision-language models are also widely applied in end-to-end autonomous driving. HE-Drive \cite{hedrive} explores the integration of vision-language models into end-to-end autonomous driving systems, enabling more human-like reasoning and the ability to handle complex driving scenarios. Similarly, DriveVLM\cite{drivevlm} further argue the integration of vision-language models with autonomous driving technology that combines visual and language information can enhance the capabilities of perception and decision-making. Additionally, DriveLM \cite{drivelm} introduces the integration of visual question answering with autonomous driving. By employing a graphical visual question answering model, the system offers a more flexible and interactive reasoning approach, enabling driving decisions to be made not only based on raw inputs but also through a question-and-answer process that further reasons and validates scene information.

Further, researchers are increasingly pursuing more integrated and generalizable models that not only handle driving tasks themselves but also address more complex perception, reasoning, and planning tasks. In Reason2Drive \cite{reason2drive}, the authors propose a chain-based reasoning framework for autonomous driving that offers more rational decision support in complex scenarios. OmniDrive \cite{omnidrive} and OpenDriveVLA\cite{opendrivevla} integrates large language models with 3D perception, reasoning, and planning, providing a more comprehensive framework that further enhances the multitasking capabilities of autonomous driving systems.

Overall, with the introduction of interpretability, multimodal learning, and large language models, research has increasingly shifted toward the development of more efficient systems with enhanced reasoning capabilities.

\section{Proposed Approach}

\noindent\textbf{Overview} 
InsightDrive comprises three sections. In Sec. 1, we present an attention-centric explicit scene representation built on a sparse instances with sparse detection and tracking, sparse online mapping, and motion prediction, which encodes interactions among the ego vehicle, surrounding agents, and the map while emphasizing planning critical regions. In Sec. 2, we introduce an implicit scene representation that models human driving cognition via Chain-of-Thought (CoT) instructions and injects this knowledge through a task-level MoE adapter, which aligns visual tokens with CoT queries and routes features to attention-centric perception, counterfactual reasoning, and planning experts for task-aware conditioning. In Sec. 3, we propose a diffusion planner conditioned on explicit attention-centric and implicit reasoning-centric representations, which produces robust and adaptive trajectories under a reconstruction and classification objective with DDIM sampling. This architecture establishes a distillation path from humans to LLMs to onboard models and unifies perception attention with cognitive reasoning for planning as show in Fig\ref{fig:framework}.

\subsection{Explicit Scene Representation}
We first construct explicit scene representation based on sparse instances, which consists of Sparse Detection $\&$ Tracking, Sparse Online Mapping, and Motion Prediction. These tasks jointly describe dynamic agents and static map elements in the driving scene.

\begin{figure}[t]
    \centering
    \includegraphics[width=0.4\textwidth]{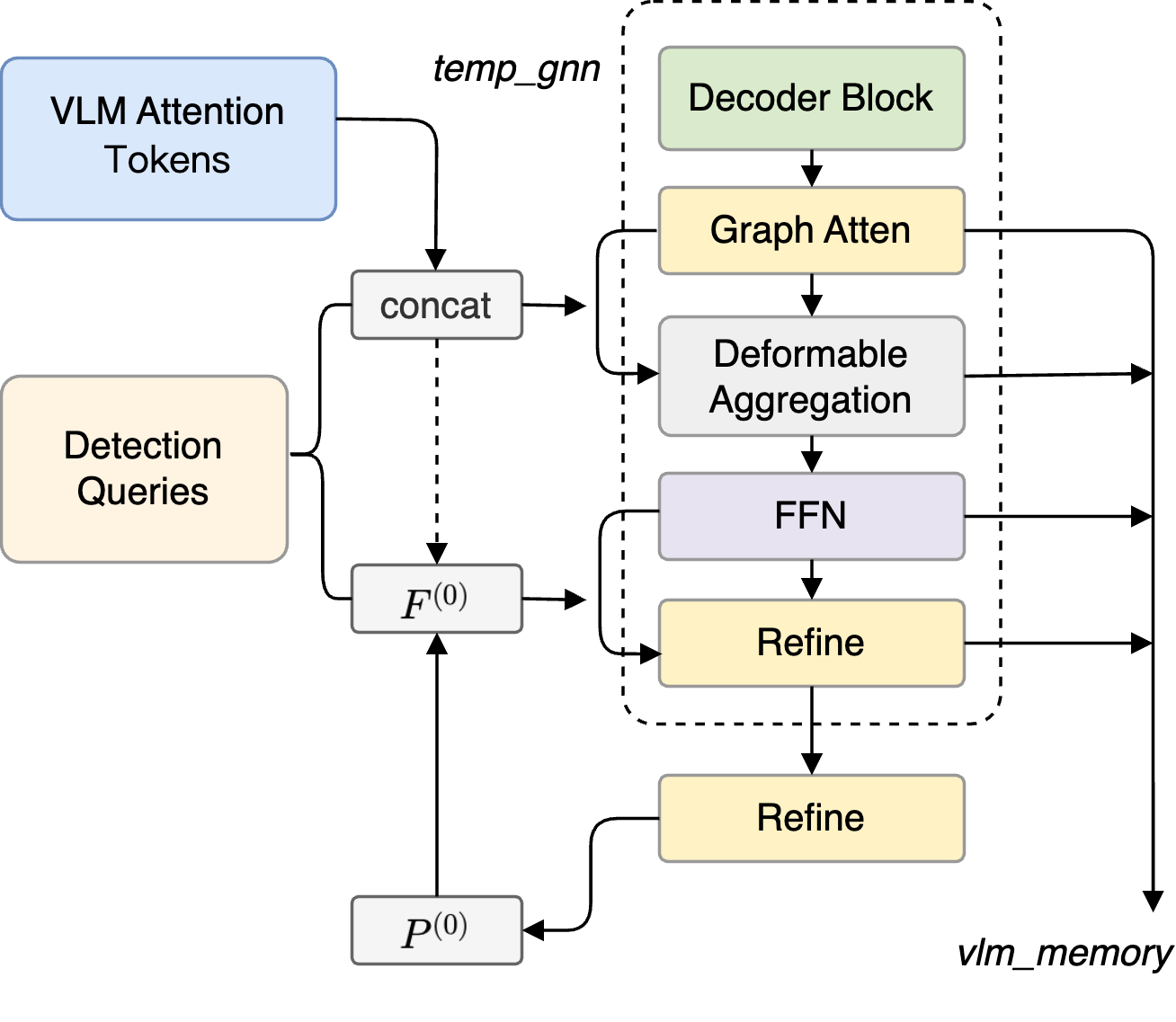} %
    \caption{The joint tokens are processed by a decoder block. The refine head outputs class scores and refined anchors $P^{(0)}$, while the corresponding query features  $F^{(0)}$ are forwarded to the next block. VLM tokens are preserved across blocks and aggregated as vlm memory for downstream modules.}
    \label{sparse}
\end{figure}

\noindent \textbf{Sparse VLM Detection.}
We extend the sparse detection and mapping head with language guidance by injecting VLM attention tokens through SparseVLMHead, which allocates $E$ learnable VLM memory tokens.
\begin{equation}
\underbrace{\mathbf{F}^{(0)}}_{\text{queries}}
= \big[\, \mathbf{F}^{(0)}_{\text{det}} \;;\; \mathbf{1}_{B}\!\otimes\!\mathbf{E}_{\text{vlm}} \,\big],
\qquad
\underbrace{\mathbf{A}^{(0)}}_{\text{anchors}}
= \big[\, \mathbf{A}^{(0)}_{\text{det}} \;;\; \mathbf{0} \,\big].
\end{equation}
Here ${F}^{(0)}_{\text{det}}$ and ${A}^{(0)}_{\text{det}}$ 
are the standard sparse detection queries and anchor states. VLM tokens use content-only embeddings , so their anchors are zero-padded to the same state dimension. Both detection and VLM anchors are encoded by a shared positional encoder:
\begin{equation}
\mathbf{P}^{(0)}=\mathrm{PE}\!\big(\mathbf{A}^{(0)}\big).
\end{equation}
The decoder acts as a query-driven iterative reasoner, in which language tokens and detection queries share a common feature space and interact via cross-attention, so that language-conditioned priors persist across layers and steer the focus toward planning-critical regions. See Fig. \ref{sparse} for architectural details.


The refine layer outputs classification scores and refined anchors for the detection queries, while the VLM tokens are sliced as memory ${M}^{(l)}$ and reintroduced before the next block, so that text-derived attention persists across decoding. The head returns detection predictions and an aggregated VLM memory for downstream modules.
\begin{equation}
    B_d = \{x, y, z, \ln w, \ln h, \ln l, \sin \theta, \cos \theta, v_x, v_y, v_z\}
\end{equation}
Surrounding agents are represented by instance features $F_d \in \mathbb{R}^{N_d \times C}$ and $\quad B_d \in \mathbb{R}^{N_d \times 11}$, where $N_d$ is the number of anchors and $C$ is the feature dimension. Each anchor box is defined by location, size, orientation, and velocity.


\noindent \textbf{Sparse VLM Mapping.}
Sparse VLM mapping follows the same pipeline as Sparse VLM detection. Injected VLM attention tokens update sparse queries for static map elements and refine their anchors, which enables language-guided extraction and vectorized regression of multi-segment polylines (e.g., lane boundaries, curbs, stop lines) under a shared positional encoder and decoder stack.

Static map elements are represented as polylines with $N_p$ points. Each element is associated with instance features and refined iteratively through the same decoder structure as detection.
It can be formulated as follows:
\begin{equation}
    l = (x_0, y_0, \dots, x_{N_p-1}, y_{N_p-1})
\end{equation}

\noindent \textbf{Motion Prediction.}
To capture agent dynamics, temporal attention is applied to fuse current and historical instance features. The motion state of each agent is represented as:
\begin{equation}
    s^t = [x^t, y^t, v_x^t, v_y^t]
\end{equation}
which serves as an explicit prior for downstream motion prediction and planning.
\subsection{Implicit Scene Representation}
While explicit scene representations capture observable geometry and semantics, robust end-to-end driving also requires cognition and reasoning ability. We therefore introduce an implicit scene representation that encodes human driving cognition and reasoning, which aligns the model with human priors and provides human-like guidance to trajectory prediction. Specifically, we model human reasoning with Chain-of-Thought (CoT) instructions and inject this knowledge into scene features via a task-level Mixture-of-Experts (MoE) adapter, which complements the explicit stream for planning.

\noindent \textbf{Chain-of-Thought for Cognition and Reasoning}
We introduce a chain-of-thought framework that simulates the cognition and reasoning process of human drivers. As illustrated in Fig.~\ref{fig:framework}(a), this process is structured into four stages: 1) Scene Description, 2) Driver Attention, 3) Factual Reasoning, and 4) Motion Planning.  Stage-aligned prompts elicit stepwise reasoning from LLMs over traffic scenarios, which produces knowledge that aligns with human driving cognition.

The CoT-generated outputs are further verified by experts to guarantee consistency with real-world driving behavior. We integrate the verified instructions into the scene representation, so that the model acquires cognitive and reasoning priors at negligible computational cost.

It establishes a human–LLM–vehicle knowledge distillation pipeline, where human cognitive strategies are first encoded into CoT instructions, then distilled into LLM outputs, and finally transferred into lightweight onboard models, which bridge the gap between human driver and end-to-end autonomous driving systems.

\noindent \textbf{Task-level Mixture-of-Experts adapter}
Our goal is to inject cognitive knowledge which is obtained through Chain-of-Thought (CoT) reasoning into the end-to-end driving model. To bridge the vision–language representational gap and to perform task-conditioned routing and expert specialization, we introduce a task-level Mixture-of-Experts (MoE) adapter, which aligns visual tokens with CoT queries and conditionally dispatches features to perception, counterfactual-reasoning, and planning experts. It consists of a vision–language alignment module that maps backbone visual tokens and CoT queries into a shared embedding space, and an MoE router that performs task-conditioned routing of the aligned features to experts for scene attention, reasoning, and planning as shown in Fig\ref{fig:framework}. This adapter effectively transfers human driving cognition into the model and provides task-aware conditioning for downstream modules with negligible parameter overhead.

\subsubsection{\textbf{Vision-Language Alignment}}

To align the visual tokens with the language queries, we employ cosine similarity as the similarity metric and compute the similarity\cite{uniter}. Next, we employ the widely used Image-Text Contrastive Loss (ITC) from contrastive learning to optimize the alignment process\cite{blip}\cite{blip2}. The loss function is defined as follows:
\begin{equation}
p^{i2t}_m(I) = \frac{\exp(s(I, T_m)/\tau)}{\sum_{m=1}^M \exp(s(I, T_m)/\tau)}, 
\end{equation}
\begin{equation}
    p^{t2i}_m(T) = \frac{\exp(s(T, I_m)/\tau)}{\sum_{m=1}^M \exp(s(T, I_m)/\tau)},
\end{equation}
where $s(I, T)$ denotes the cosine similarity between the image embedding $I$, and the text embedding $T$. $\tau$ is a temperature parameter controlling the sharpness of the similarity distribution. The subscript $m$ indexes candidate pairs in the batch, with the positive pair corresponding to the one-hot label 
$y$.

The final text–image contrastive loss is expressed as the symmetric cross entropy:
\begin{equation}
\begin{aligned}
\mathcal{L}_{itc} = \tfrac{1}{2}\,\mathbb{E}_{(I,T)\sim D}\Big[
   & H(y^{i2t}(I), p^{i2t}(I)) \\
 + & H(y^{t2i}(T), p^{t2i}(T))
\Big],
\end{aligned}
\end{equation}
where $H(.,.)$ denotes the cross-entropy function. This objective encourages the model to learn a robust joint vision–language embedding that captures CoT knowledge for downstream planning.

In addition, we employ Image-grounded Text Generation (ITG) method to transform image features into driving-related textual descriptions. 
Given a feature representation $\mathbf{A} \in \mathbb{R}^{M \times d}$, where $M$ denotes the number of vision feature tokens and $d$ represents the feature dimension, the model aims to learn the conditional probability:
\begin{equation}
    P(\mathbf{T} \mid \mathbf{A}) = P(t_1, t_2, ..., t_N \mid \mathbf{A})
\end{equation}
where $\mathbf{T} = (t_1, t_2, ..., t_N)$ represents the target text sequence consisting of $N$ discrete tokens. The text generation process follows an autoregressive manner, formulated as:
\begin{equation}
    P(\mathbf{T} \mid \mathbf{A}) = \prod_{i=1}^{N} P(t_i \mid t_{<i}, \mathbf{A})
\end{equation}
where $P(t_i \mid t_{<i}, \mathbf{A})$ denotes the probability of predicting the current token $t_i$ given all previous tokens $t_{<i}$ and the vision feature representation $\mathbf{A}$.

The textual representation is obtained using a pre-trained language model\cite{bert}. The model is trained to minimize the negative log-likelihood of the correct token sequence using cross-entropy loss:
\begin{equation}
    L_{\text{ITG}} = - \sum_{i=1}^{N} \log P(t_i \mid t_{<i}, \mathbf{A})
\end{equation}
where $P(t_i \mid t_{<i}, \mathbf{A})$ represents the probability of predicting token $t_i$ based on previous tokens and the visual tokens. The objective is to maximize the likelihood of generating accurate and contextually relevant descriptions.

\subsubsection{\textbf{Task-level Mixture-of-Experts}}
To inject cognitive knowledge derived from Chain-of-Thought (CoT) into the driving model, we introduce a Task-level Mixture-of-Experts. The aligned vision–language emmbedings $\hat{F}$ is routed to different experts through an MoE router:
\begin{equation}
g = \text{Softmax}(W_r \hat{F})
\end{equation}
\begin{equation}
F_{\text{moe}} = \sum_{k=1}^{K} g_k \cdot \text{Expert}_k(\hat{F}),
\end{equation}
where $W_r$ is the routing projection, $g_k$
is the gating weight of expert k, and $expert_k$ denotes a task-specific expert network.

CoT-derived knowledge is routed to three expert groups—perception, counterfactual reasoning, and planning—each aligned with a distinct driving sub-task. We optimize the model end to end with a multi-task objective:
\begin{equation}
{L}_{\text{total}} =
\lambda_{\text{per}} {L}_{\text{per}} +
\lambda_{\text{ctr}} {L}_{\text{ctr}} +
\lambda_{\text{plan}} {L}_{\text{plan}},
\end{equation}
Perception experts produce features that are fed into Sparse Detection and Sparse Online Mapping to refine object and map representations while directing perception attention. Counterfactual reasoning experts generate features that encode alternative intentions and are injected into the planning module to improve robustness in uncertain or ambiguous scenarios. Planning experts output features that directly support trajectory generation while offering planning-aware conditioning for decision making.

\subsection{Diffsuion Planner}
Human driving behaviors are shaped by structured cognition and contextual reasoning, rather than unconstrained denoising from random noise as in vanilla diffusion policies. To align trajectory generation with human driving cognition, we propose a conditioned diffusion planner that generates trajectories under the joint guidance of attention-centric explicit scene representations and reasoning-based implicit scene representations.

\noindent \textbf{Training.} Given a ground-truth trajectory $\tau_{\ast} = \{(x_t, y_t)\}_{t=1}^{T}$, we define the forward diffusion process as:
\begin{equation}
\tilde{\tau}_i = \sqrt{\gamma_i}\,\tau_{\ast} + \sqrt{1 - \gamma_i}\,\epsilon, 
\quad \epsilon \sim {N}(0, I),
\end{equation}
where $i \in [1, T]$ is the diffusion step and $\gamma_i$ is the noise scheduling coefficient. At each step, the diffusion decoder $f_\theta$ takes as input the noisy trajectory $\hat{\tau_i}$ together with a composite condition vector:
\begin{equation}
\{\hat{s}, \hat{\tau}\} = f_\theta(\tilde{\tau}_i, \mathbf{C}),
\end{equation}
where $\mathbf{C} = [F^{\text{exp}}, F^{\text{imp}}]$ encodes both explicit attention-based features and implicit reasoning features. The decoder outputs denoised trajectories $\hat{\tau}$ and auxiliary classification scores $\hat{s}$.

The training objective combines trajectory reconstruction and trajectory-level classification:
\begin{equation}
{L} = {L}_{\text{rec}}(\hat{\tau}, \tau_{\ast}) + 
\lambda \,\text{BCE}(\hat{s}, y),
\end{equation}
where $L_{rec}$ is an $L1$ reconstruction loss, 
$BCE$ denotes binary cross-entropy, and $\lambda$ balances the two terms.

\textbf{Inference.}
At inference, we initialize trajectories by sampling $\tilde{\tau}_T \sim \mathcal{N}(0, I)$ and iteratively denoise them into planning outputs. At step $i$, the decoder refines $\tilde{\tau}_T$ conditioned on $C$, so that explicit attention and implicit reasoning jointly guide generation. We employ a DDIM sampler for efficient sampling, which keeps the predicted trajectories aligned with human driving cognition.

\newcommand{\arraysep}{\renewcommand\arraystretch{1.2}}

\begin{table*}[t]
\setlength{\tabcolsep}{0.01\linewidth}
\caption{\textbf{Comparisons results with latest SOTA methods on the NuScenes\cite{nuscenes} val dataset.} The ego status was not used in planning module.
We use bold to represent the best results. The comparison results are taken directly from the original paper. $\dagger$ denotes methods based on large language models (LLMs).
}
\centering
\fontsize{12}{14}\selectfont
\resizebox{0.9\textwidth}{!}{
\begin{tabular}{l|cc|ccc>{\columncolor{gray!30}}l|ccc>{\columncolor{gray!30}}l|cc}
\toprule
\multirow{2}{*}{Method} & \multirow{2}{*}{Input}  & \multirow{2}{*}{Scene Representation} &
\multicolumn{4}{c|}{L2 (m) $\downarrow$} & 
\multicolumn{4}{c|}{Collision Rate (\%) $\downarrow$} &
\\
& & &  1s & 2s & 3s & Avg. & 1s & 2s & 3s & Avg. & \multirow{-2}*{LLM} & \multirow{-2}*{FPS} \\
\midrule
UniAD~\cite{uniad} & C & Det \& Map \&  Occ \& Motion & 0.44 & 0.67 & 0.96 & 0.69 & 0.04 & 0.08 & 0.23 & 0.12 & w/o & 1.8 \\
VAD~\cite{vad} & C & Det \& Map \& Motion & 0.41 & 0.70 & 1.05 & 0.72 & 0.07 & 0.17 & 0.41 & 0.22 & w/o & 4.5 \\
BEV-Planner~\cite{ego}& C & None & 0.28 & 0.42 & 0.68 & 0.46 & 0.04 & 0.37 & 1.07 & 0.49 & w/o & - \\
PARA-Drive~\cite{paradrive} & C & Det \& Map \& Motion & 0.25 & 0.46 & 0.74 & 0.48 & 0.14 & 0.23 & 0.39 & 0.25 & w/o & 5.0 \\
GenAD~\cite{genad} & C & Det \& Map \& Motion & 0.28 & 0.58 & 0.96 & 0.52 & 0.08 & 0.14 & 0.34 & 0.19 & w/o & 4.2 \\
OccWorld~\cite{occworld} & C & Occ & 0.39 & 0.73 & 1.18 & 0.77 & 0.11 & 0.19 & 0.67 & 0.32 & w/o & 2.8 \\
SparseDrive~\cite{SparseDriveEA} & C & Det \& Map \& Motion & 0.29 & 0.58 & 0.96 & 0.61 & 0.05 & 0.18 & 0.34 & 0.18 & w/o & 9.0 \\
DiffusionDrive~\cite{diffusiondrive} & C  & Det \& Map \& Motion & 0.27 & 0.54 & 0.90 & 0.57 & 0.03 & 0.05 & 0.16 & 0.08 & w/o & 8.2 \\
\midrule
DriveVLM$^\dagger$~\cite{drivevlm} & C & Det \& Motion \& Imp & 0.18 & 0.34 & 0.68 & 0.40 & 0.01 & 0.22 & 0.45 & 0.27 & Qwen-VL-7B & - \\
GPT-Driver$^\dagger$~\cite{gpt-driver} & C & Det \& Motion \& Imp & 0.20 & 0.40 & 0.70 & 0.44 & 0.04 & 0.12 & 0.36 & 0.17 & GPT-3.5 & - \\
RDA-Driver$^\dagger$~\cite{rda} & C & Det \& Motion \& Imp & 0.17 & 0.37 & 0.69 & 0.40 & 0.01 & 0.05 & 0.26 & 0.10 &LLaVa-7B & - \\
OpenDriveVLA$^\dagger$~\cite{opendrivevla} & C & Det \& Map \& Motion \& Imp & \textbf{0.15} & \textbf{0.31} & \textbf{0.55} & \textbf{0.33} & 0.01 & 0.08 & 0.21 & 0.10 & Qwen2.5-7B & - \\
\midrule
\textbf{InsightDrive(ours)} & C & Det \& Map \& Motion \& Imp &  0.26 &  0.53 & 0.90 & 0.56  & \textbf{0.00} & \textbf{0.01} & \textbf{0.09} & \textbf{0.03} & w/o & 7.8 \\
\bottomrule
\end{tabular}%
}
\label{tab:nuscenes}
\end{table*}

\begin{table*}[t]
\setlength{\tabcolsep}{0.01\linewidth}
\caption{\textbf{Closed-loop comparison on the NAVSIM benchmark.}
We use \textbf{bold} for the best. }
\centering
\fontsize{12}{14}\selectfont
\resizebox{0.9\textwidth}{!}{
\begin{tabular}{l|c c|c c c c c|>{\columncolor{gray!30}}c}
\toprule
Method & Input & Scene Representation &
NC$\uparrow$ & DAC$\uparrow$ & TTC$\uparrow$ & Comf.$\uparrow$ & EP$\uparrow$ &
PDMS$\uparrow$ \\
\midrule
Constant Velocity & - & - & 68.0 & 57.8 & 50.0 & \textbf{100} & 19.4 & 20.6 \\
Ego Status MLP & - & -  & 93.0 & 77.3 & 83.6 & \textbf{100} & 62.8 & 65.6 \\
\midrule
Hydra-MDP-$\mathcal{V}_{8192}$~\cite{hydra} & C \& L & Det \& Map \& Motion & 97.9 & 91.7 & 92.9 & \textbf{100} & 77.6 & 83.0 \\
UniAD~\cite{uniad} & C & Det \& Map \& Occ \& Motion  & 97.8 & 91.9 & 92.9 & \textbf{100} & 78.8 & 83.4 \\
VADv2-$\mathcal{V}_{8192}$~\cite{vadv2} & C & Det \& Map \& Motion & 97.2 & 89.1 & 91.6 & \textbf{100} & 76.0 & 80.9 \\
LTF~\cite{transfuser}                    & C & Det \& Map \& Occ    & 97.4 & 92.8 & 92.4 & \textbf{100} & 79.0 & 83.8 \\
Transfuser~\cite{transfuser}            & C \& L & Det \& Map \& Seg \& Depth & 97.7 & 92.8 & 92.8 & \textbf{100} & 79.2 & 84.0 \\
PARA-Drive~\cite{paradrive}      & C & Map \& Occ \& Motion   & 97.9 & 92.4 & 93.0 & 99.8  & 79.3 & 84.0 \\

DRAMA~\cite{drama}               & C \& L & Det \& Map & 98.0 & 93.1 & 94.8 & \textbf{100} & 80.1 & 85.5 \\

Hydra-MDP-$\mathcal{V}_{8192}$-W-EP~\cite{hydra} & C \& L & Det \& Map \& Motion & 98.3 & 96.0 & 94.6 & \textbf{100} & 78.7 & 86.5 \\

DiffusionDrive~\cite{diffusiondrive}   & C \& L & Det \& Map  & 98.2 & 96.2 & 94.7 & \textbf{100} & 82.2 & 88.1 \\

WoTE~\cite{wote}   & C \& L & Det \& Map  & \textbf{98.5} & \textbf{96.8} & 94.9 & 99.9 & 81.9 & 88.3 \\
\midrule
\textbf{InsightDrive(ours)}   & C \& L & Det \& Map \& Imp  & \textbf{98.5} & 96.7 & \textbf{95.2} & 99.9 & \textbf{82.3} & \textbf{88.5} \\
\bottomrule
\end{tabular}
}
\label{tab:navsim}
\end{table*}

\section{Experiments}

\begin{table*}[t]
\centering
\setlength{\tabcolsep}{6pt}
\caption{\textbf{Comparisons of perception performance.} 
L $\rightarrow$ A represents our language attention guidance approach to guide the driver’s attention to regions that impact driving.
}
\label{tab:nuscenes_perception}
\resizebox{0.8\linewidth}{!}{
\begin{tabular}{l l c c c c c c >{\columncolor{gray!20}}c}
\toprule
Setting & Backbone & mAP$\uparrow$ & mATE$\downarrow$ & mASE$\downarrow$ & mAOE$\downarrow$ & mAVE$\downarrow$ & mAAE$\downarrow$ & NDS$\uparrow$ \\
\midrule
w/o L $\rightarrow$ A & ResNet50  & 0.412 & \textbf{0.566} & 0.276 & \textbf{0.529} & \textbf{0.265} & 0.194 & \textbf{0.522} \\
w/ L $\rightarrow$ A & ResNet50 & 0.412 & 0.587 & \textbf{0.274} & 0.539 & 0.269 & \textbf{0.191} & 0.520 \\
\bottomrule
\end{tabular}}
\label{tab:perception}
\end{table*}



\subsection{Datasets and Evaluation Metrics}
 

\begin{figure*}[t]
\centering 
\includegraphics[width=0.98\textwidth]{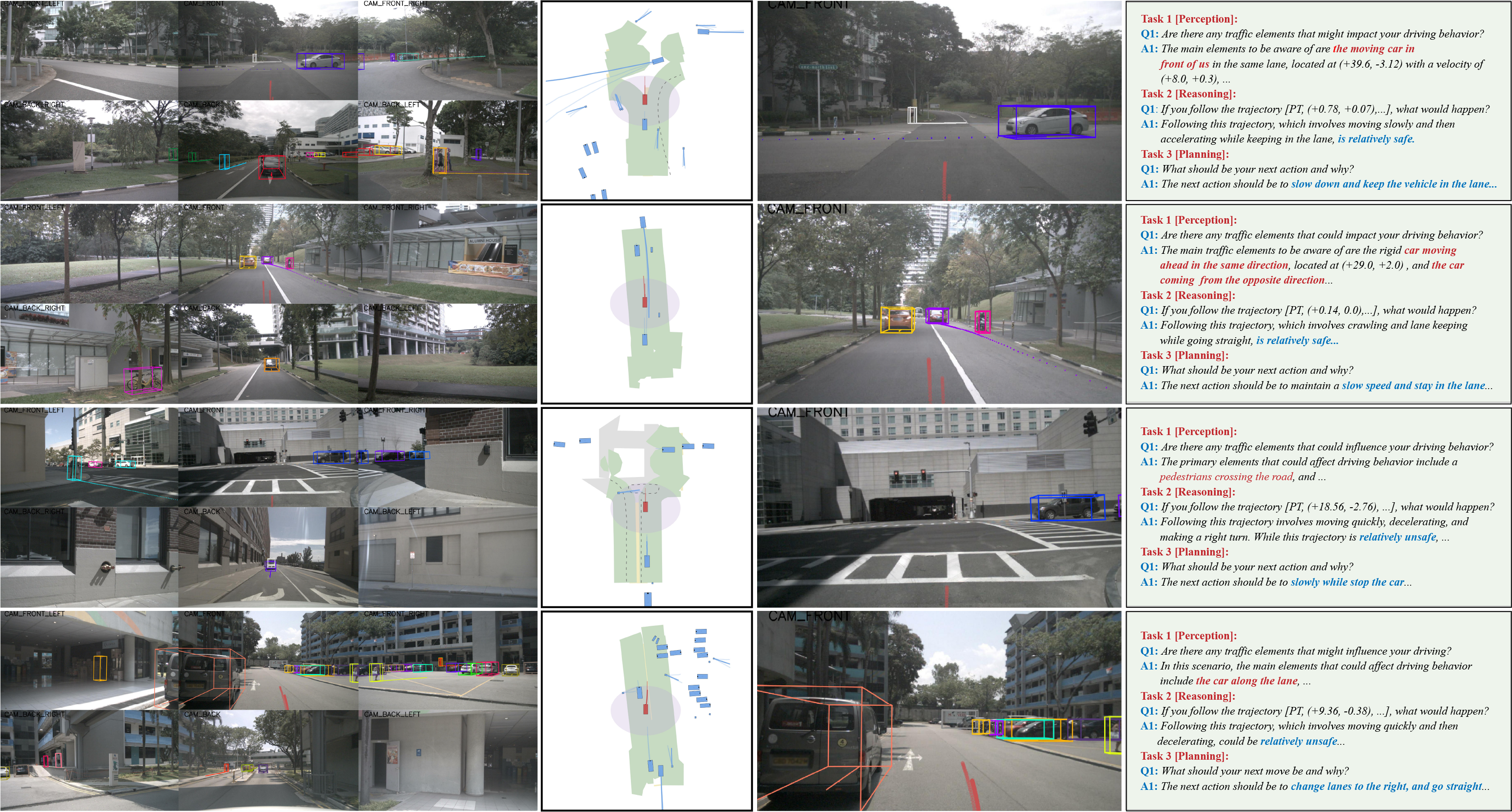}
\caption{\textbf{Visualization results of InsightDrive in several scenarios.}
We observe that InsightDrive performs well on challenging scenarios.
}
\label{fig:vis}
\end{figure*}

\textbf{Datasets} We evaluate the proposed InsightDrive framework on both the NuScenes\cite{nuscenes} and the NAVSIM\cite{navsim} benchmark.
The NuScenes dataset contains 1000 driving scenes of 20 seconds each, captured with six surround-view cameras and a 32-beam LiDAR. Following the official split, we use 700 scenes for training, 150 for validation, and 150 for testing.
To complement open-loop evaluation on NuScenes, we adopt NAVSIM\cite{navsim}, a non-reactive simulation benchmark, that unrolls predicted trajectories across diverse scenarios and evaluates progress, safety, and comfort.

\textbf{Evaluation Metrics} Following existing end-to-end autonomous driving methods, we adopt the L2 displacement error and collision rate for open-loop evaluation.
For closed-loop evaluation on On Navsim, we follow the official metrics: No At-Fault Collision (NC↑), Drivable Area Compliance (DAC↑), Time-to-Collision (TTC↑), Comfort (Comf.↑), and Ego Progress (EP↑). These are aggregated into the Predictive Driver Model Score (PDMS↑), which serves as the overall closed-loop score.

\subsection{Implementation Details.}
The experiments use ResNet34\cite{resnet} as the backbone network to extract image features with the official pre-trained weights. In the first phase, the vision-language alignment module is pre-trained on the NuScenes dataset, while in the second phase, the end-to-end model is trained using pre-trained weights.
The model is trained for 99 epochs with a batch size of 64 on 8 NVIDIA L20 GPUs, utilizing the AdamW optimizer and a cosine annealing learning rate schedule. 

\subsection{Comparisons with State-of-the-art Methods}

\noindent \textbf{Open Loop Results}
We conducted comparative evaluations against recent end-to-end frameworks. As shown in Table~\ref{tab:nuscenes}, LLM-based approaches, which benefit from strong fine-tuning capacity, achieve lower L2 error than traditional end-to-end baselines. However, the absence of explicit scene representation correlates with higher collision rates. In contrast, InsightDrive employs a joint explicit and implicit scene representation. It achieves the lowest collision rate at real-time inference speed, which indicates safer planning. We attribute the performance to the incorporation of CoT-based cognitive knowledge, which injects attention-centric and reasoning-centric priors so that trajectory selection is interaction-aware and consistent with scene context.

\noindent \textbf{Closed Loop Results} On the Navsim benchmark, we compare against recent end-to-end frameworks. As shown in Table~\ref{tab:navsim}, InsightDrive attains the highest PDMS, with strong NC, DAC, TTC, Comfort, and Ego Progress. Relative to DiffusionDrive~\cite{diffusiondrive},it achieves a lower collision rate and a higher PDMS, indicating safer, more reliable planning under interactive rollouts. In closed-loop training, we freeze the vision–language adapter learned in open loop and optimize the remaining modules, which maintains CoT-based feature extraction capability.

\begin{table}[t]
\setlength{\tabcolsep}{0.021\linewidth}
\caption{\textbf{Effect of the insight scene representation for end-to-end autonomous driving.} 
SA and RP denote \textbf{S}cene \textbf{A}ttention and \textbf{R}easoning \textbf{P}lanning, respectively.
}
\resizebox{0.48\textwidth}{!}{
\begin{tabular}{cc|cccc|cccc}
\toprule
\multirow{2}{*}{SA} & \multirow{2}{*}{RP} &
\multicolumn{4}{c|}{L2 (m) $\downarrow$} & 
\multicolumn{4}{c}{Collision Rate (\%) $\downarrow$}  \\
& & 1s & 2s & 3s & \cellcolor{gray!30}Avg. & 1s & 2s & 3s & \cellcolor{gray!30}Avg.  \\
\midrule
$\times$ & $\times$ & 0.27 & 0.54 & 0.90 & \cellcolor{gray!30}0.57 & 0.03 & 0.05 & 0.16 & \cellcolor{gray!30}0.08	  \\

$\checkmark$ & $\times$ & 0.27 & 0.53 & 0.90  & \cellcolor{gray!30}0.57 & 0.01 & 0.03 & 0.12 & \cellcolor{gray!30}0.05  \\

$\times$ & $\checkmark$ & 0.26 & 0.53 & 0.90  & \cellcolor{gray!30}0.57 & 0.02 & 0.04 & 0.11 & \cellcolor{gray!30}0.06  \\

$\checkmark$ & $\checkmark$ & \textbf{0.26} & \textbf{0.53} & \textbf{0.90} & \cellcolor{gray!30}\textbf{0.56} & \textbf{0.00} & \textbf{0.01} & \textbf{0.09} & \cellcolor{gray!30}\textbf{0.03}  \\
\bottomrule
\end{tabular}%
}
\label{tab:abl_nuscenes}
\end{table}

\subsection{Ablation Study}
\noindent\textbf{Effect of insight scene representation in open-loop.} 
We analyze the contribution of two components of the Insight representation, which are attention-centric scene representation(SA) and implicit reasoning Planning(RP). On nuScenes, it reduces L2 displacement error and lowers collision incidence as show in Table\ref{tab:abl_nuscenes}.

\noindent\textbf{Effect of insight scene representation in closed-loop.} 
In Navsim\cite{navsim} closed-loop evaluations, diffusion-based planner improves NC, DAC, TTC, EP, and PDMS by iteratively denoising trajectories, prioritizing collision-free options, and adapting to dynamic scene contexts. By integrating chain-of-thought (CoT) reasoning, which focuses on planning-critical regions and suppresses irrelevant context (Table \ref{tab:abl_navsim}) while inferring latent intent and multi-agent interaction cues beyond pure perception, our approach achieves the lowest collision rate and the highest PDMS, outperforming all baselines. For closed-loop experiments, we freeze the vision–language adapter trained in open loop to preserve reasoning features and stabilize rollouts.

\noindent\textbf{Perception performance.} 
We further explore the necessity of the attention-centric explicit scene representation. Although global perception achieves higher perception performance, as shown in Table\ref{tab:nuscenes_perception}. It improves perception metrics worsens the collision rate as shown in Table\ref{tab:abl_navsim}.This indicates that global perception is not strictly required for end-to-end driving. Rather, an attention-centric explicit scene representation better aligns with the planning objective.

\begin{table}[t]
\setlength{\tabcolsep}{0.01\linewidth}
\caption{\textbf{Ablation on components of InsightDrive on the Navsim closed-loop benchmark.}
We present the impacts of the diffusion planner and chain-of-thought (CoT) reasoning on the performance of model in closed-loop evaluation.}
\centering
\fontsize{12}{14}\selectfont
\resizebox{0.48\textwidth}{!}{
\begin{tabular}{c|c c|ccccc|>{\columncolor{gray!30}}c}
\toprule
ID & Diffusion  & CoT
& \multicolumn{6}{c}{Planning Metric}  \\
& Planner& Reasoning& NC$\uparrow$ & DAC$\uparrow$ & TTC$\uparrow$ & Comf.$\uparrow$ & EP$\uparrow$ & PDMS$\uparrow$\\
\midrule
1 & $\times$ & $\times$ & 97.7 & 92.8 & 92.8 & 100 & 79.2 & 84.0 \\
2 & $\checkmark$ & $\times$ & 98.1 & 95.8 & 94.4 & 99.9 & 81.9 & 87.5 \\
3 & $\checkmark$ & $\checkmark$ & 98.5 & 96.7 & 95.2 & 99.9 & 82.3 & 88.5 \\
\bottomrule
\end{tabular}
}
\label{tab:abl_navsim}
\end{table}

\subsection{Visualizations.} 
We perform a visualization of InsightDrive as shown in Fig\ref{fig:vis}. We presented multiple aspects, including map segmentation, detection, motion prediction, planning results, and cot description, along with surround-view camera inputs. We observe that InsightDrive achieves excellent performance in extracting key information, which has a significant impact on driving. It effectively reduces the risk of collisions. Experimental results demonstrate that InsightDrive achieves more precise and safer trajectory planning in several driving scenarios, such as emergency braking, lane chang, crosswalk yielding and lane borrowing.

\section{Conclusion}



In this paper, we present InsightDrive, a lightweight end-to-end framework that unifies an attention-centric explicit scene representation with a reasoning-centric implicit scene representation. Human driving cognition is captured via CoT instructions and injected through a task-level MoE adapter that performs task-conditioned routing to perception, counterfactual-reasoning, and planning experts. A diffusion planner conditioned on a joint explicit–implicit scene representation generates robust, adaptive trajectories. Experiments demonstrate robust improvements on both NuScenes and Navsim benchmark.
In the future, we will further explore scene representation methods suitable for end-to-end tasks and work toward scalable real-world deployment.


\bibliographystyle{IEEEtran}
\bibliography{egbib}

\end{document}